\pdfoutput=1
\documentclass[11pt,dvipsnames]{article}

\usepackage{acl}

\usepackage{times}
\usepackage{latexsym}

\usepackage[T1]{fontenc}
\usepackage[utf8]{inputenc}
\usepackage{times}
\usepackage{latexsym}
\usepackage{url}
\usepackage{enumitem}
\usepackage{amsmath,amssymb,amsfonts}
\usepackage{algorithmic}
\usepackage{caption}
\usepackage{subcaption}
\usepackage{graphicx}
\usepackage{textcomp}
\usepackage{url}
\usepackage{makecell}
\usepackage{multirow}
\usepackage{booktabs}
\usepackage{array}
\usepackage{microtype}

\usepackage{multirow}
\usepackage{booktabs}

\usepackage{hyperref}       % hyperlinks
\usepackage{amsfonts}       % blackboard math symbols
\usepackage{nicefrac}       % compact symbols for 1/2, etc.

\newcommand{\thickhline}{%
    \noalign {\ifnum 0=`}\fi \hrule height 1pt
    \futurelet \reserved@a \@xhline
}

\usepackage{pifont}
\newcommand{\cmark}{O}%
\newcommand{\xmark}{\ding{55}}%
\definecolor{lp}{HTML}{CBC3E3}
\usepackage{colortbl}
\usepackage{pythonhighlight}
\usepackage{xspace}
\newcommand{\pt}{SOLAR 10.7B\xspace}
\newcommand{\ft}{SOLAR 10.7B-Instruct\xspace}

\usepackage{comment}
\usepackage{todonotes}
\usepackage{microtype}
\usepackage[cjk]{kotex}
\newcommand\blfootnote[1]{%
  \begingroup
  \renewcommand\thefootnote{}\footnote{#1}%
  \addtocounter{footnote}{-1}%
  \endgroup
}

\title{SOLAR 10.7B: Scaling Large Language Models with Simple yet Effective Depth Up-Scaling}

\author{Dahyun Kim$^{*}$, Chanjun Park$^{*\dagger}$, Sanghoon Kim$^{*\dagger}$, Wonsung Lee$^{*\dagger}$, Wonho Song$^{*}$  \\ {\bf \large Yunsu Kim$^{*}$, Hyeonwoo Kim$^{*}$, Yungi Kim, Hyeonju Lee, Jihoo Kim}\\ {\bf \large  Changbae Ahn, Seonghoon Yang, Sukyung Lee, Hyunbyung Park, Gyoungjin Gim}\\{\bf \large Mikyoung Cha, Hwalsuk Lee$^{\dagger}$, Sunghun Kim$^{\dagger}$}\\ \\
  Upstage AI, South Korea \\
  \texttt{\footnotesize\{kdahyun, chanjun.park, limerobot, wonsung.lee, hwalsuk.lee, hunkim\}@upstage.ai}}

\begin{document}
\maketitle
\begin{abstract}
\blfootnote{$^*$Equal Contribution $^\dagger$ Corresponding Author}
We introduce \pt, a large language model (LLM) with 10.7 billion parameters, demonstrating superior performance in various natural language processing (NLP) tasks. Inspired by recent efforts to efficiently up-scale LLMs, we present a method for scaling LLMs called depth up-scaling (DUS), which encompasses depthwise scaling and continued pretraining. In contrast to other LLM up-scaling methods that use mixture-of-experts, DUS does not require complex changes to train and inference efficiently. We show experimentally that DUS is simple yet effective in scaling up high-performance LLMs from small ones. Building on the DUS model, we additionally present \ft, a variant fine-tuned for instruction-following capabilities, surpassing Mixtral-8x7B-Instruct. \pt is publicly available under the Apache 2.0 license, promoting broad access and application in the LLM field~\footnote{\url{https://huggingface.co/upstage/SOLAR-10.7B-v1.0}}.
\end{abstract}

\section{Introduction}
\begin{figure*}[t!]
    \centering
    \resizebox{0.80\linewidth}{!}{
    \includegraphics{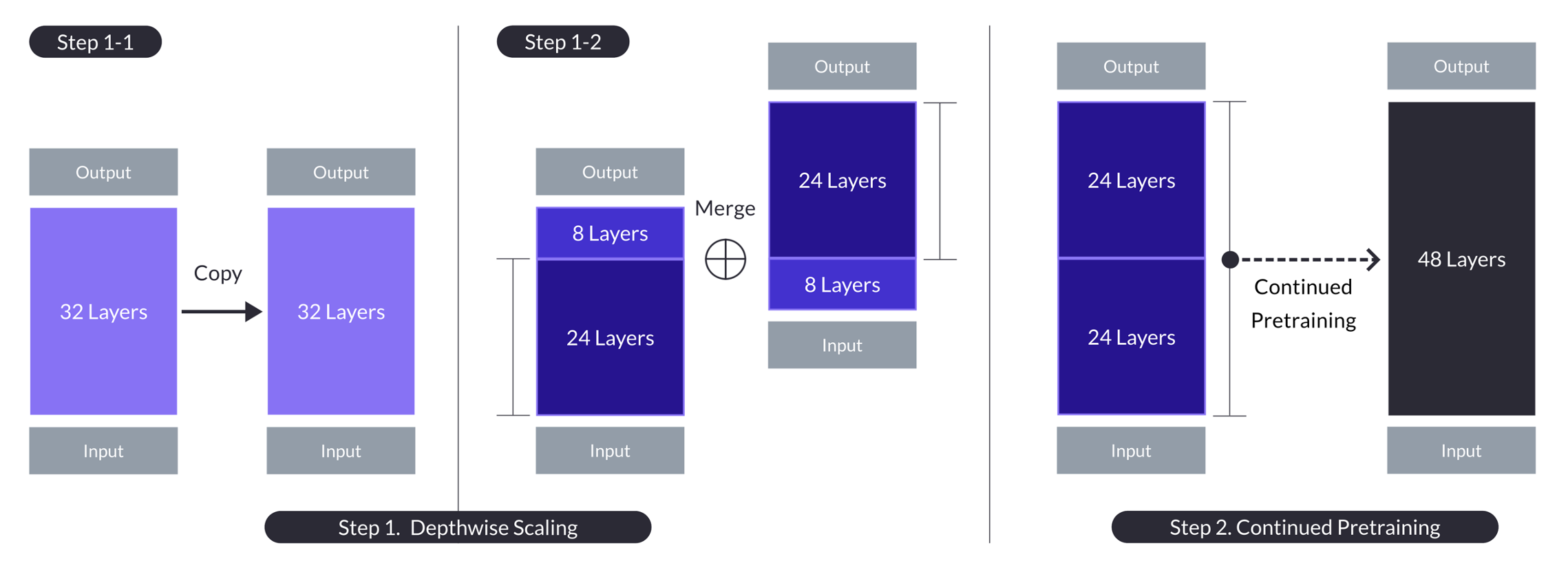}
    }
    \caption{Depth up-scaling for the case with $n=32, s=48,$ and $m=8$. Depth up-scaling is achieved through a dual-stage process of depthwise scaling followed by continued pretraining.}
    \label{fig:dus}
\end{figure*}

The field of natural language processing (NLP) has been significantly transformed by the introduction of large language models (LLMs), which have enhanced our understanding and interaction with human language~\cite{zhao2023survey}. These advancements bring challenges such as the increased need to train ever larger models~\cite{rae2021scaling,wang2023learning,pan2023reusing,winglian,yao20232x,gesmundo2023composable} owing to the performance scaling law~\cite{kaplan2020scaling, hernandez2021scaling, anil2023palm,kaddour2023no}. To efficiently tackle the above, recent works in scaling language models such as a mixture of experts (MoE)~\cite{shazeer2017outrageously,komatsuzaki2022sparse} have been proposed. While those approaches are able to efficiently and effectively scale-up LLMs, they often require non-trivial changes to the training and inference framework~\cite{gale2023megablocks}, which hinders widespread applicability. Effectively and efficiently scaling up LLMs whilst also retaining the \textit{simplicity} for ease of use is an important problem~\cite{alberts2023large,fraiwan2023review,sallam2023chatgpt,bahrini2023chatgpt}.

Inspired by ~\citet{komatsuzaki2022sparse}, we present depth up-scaling (DUS), an effective and efficient method to up-scale LLMs whilst also remaining straightforward to use.
DUS consists of scaling the number of layers in the base model and continually pretraining the scaled model. Unlike \cite{komatsuzaki2022sparse}, DUS does not scale the model using MoE and rather use a depthwise scaling method analogous to ~\citet{tan2019efficientnet} which is adapted for the LLM architecture. Thus, there are no additional modules or dynamism as with MoE, making DUS immediately compatible with easy-to-use LLM frameworks such as HuggingFace~\cite{wolf2019huggingface} with no changes to the training or inference framework for maximal efficiency. Furthermore, DUS is applicable to all transformer architectures, opening up new gateways to effectively and efficiently scale-up LLMs in a simple manner. Using DUS, we release \pt, an LLM with 10.7 billion parameters, that outperforms existing models like Llama 2~\cite{touvron2023llama2} and Mistral 7B~\cite{jiang2023mistral} in various benchmarks.

We have also developed \ft, a variant fine-tuned for tasks requiring strict adherence to complex instructions. It significantly outperforms the Mixtral-8x7B-Instruct model across various evaluation metrics, evidencing an advanced proficiency that exceeds the capabilities of even larger models in terms of benchmark performance.

By releasing \pt under the Apache 2.0 license, we aim to promote collaboration and innovation in NLP. This open-source approach allows for wider access and application of these models by researchers and developers globally.

\section{Depth Up-Scaling}
To efficiently scale-up LLMs, we aim to utilize pretrained weights of base models to scale up to larger LLMs~\cite{komatsuzaki2022sparse}. While existing methods such as \citet{komatsuzaki2022sparse} use MoE~\cite{shazeer2017outrageously} to scale-up the model architecture, we opt for a different depthwise scaling strategy inspired by \citet{tan2019efficientnet}. 
We then continually pretrain the scaled model as just scaling the model without further pretraining degrades the performance.

\begin{table*}[t!]
    \centering
    \resizebox{0.99\linewidth}{!}{
\begin{tabular}{ccccccc}
\toprule
 \multirow{3}{*}{Properties}   & \multicolumn{6}{c}{Training Datasets}\\
 &  \multicolumn{3}{c}{Instruction} & \multicolumn{3}{c}{Alignment} \\ \cmidrule(lr){2-4} \cmidrule(lr){5-7}
& Alpaca-GPT4 & OpenOrca & Synth. Math-Instruct & Orca DPO Pairs & Ultrafeedback Cleaned & Synth. Math-Alignment \\ \cmidrule(lr){1-1} \cmidrule(lr){2-4} \cmidrule(lr){5-7}
Total \# Samples &52K & 2.91M & 126K & 12.9K & 60.8K & 126K\\
Maximum \# Samples Used &52K & 100K & 52K & 12.9K & 60.8K & 20.1K\\
Open Source & {\color{ForestGreen} \cmark}& {\color{ForestGreen} \cmark} & {\color{red} \xmark} & {\color{ForestGreen} \cmark} & {\color{ForestGreen} \cmark} & {\color{red} \xmark}\\
\bottomrule
\end{tabular}
}
\caption{Training datasets used for the instruction and alignment tuning stages, respectively. For the instruction tuning process, we utilized the Alpaca-GPT4~\cite{peng2023instruction}, OpenOrca~\cite{mukherjee2023orca}, and Synth. Math-Instruct datasets, while for the alignment tuning, we employed the Orca DPO Pairs~\cite{intel2023orcadpo}, Ultrafeedback Cleaned~\cite{cui2023ultrafeedback,ivison2023camels}, and Synth. Math-Alignment datasets. The `Total \# Samples` indicates the total number of samples in the entire dataset. The `Maximum \# Samples Used` indicates the actual maximum number of samples that were used in training, which could be lower than the total number of samples in a given dataset. `Open Source` indicates whether the dataset is open-sourced.}
\label{tab:datasets}
\end{table*}

\paragraph{Base model.}
Any $n$-layer transformer architecture can be used but we select the 32-layer Llama 2 architecture as our base model. We initialize the Llama 2 architecture with pretrained weights from Mistral 7B, as it is one of the top performers compatible with the Llama 2 architecture.
By adopting the Llama 2 architecture for our base model, we aim to leverage the vast pool of community resources while introducing novel modifications to further enhance its capabilities. 

\paragraph{Depthwise scaling.}
\label{para:dus}

From the base model with $n$ layers, we set the target layer count $s$ for the scaled model, which is largely dictated by the available hardware. 

With the above, the depthwise scaling process is as follows. The base model with $n$ layers is duplicated for subsequent modification. Then, we remove the final $m$ layers from the original model and the initial $m$ layers from its duplicate, thus forming two distinct models with $n-m$ layers. These two models are concatenated to form a scaled model with $s = 2 \cdot (n-m)$ layers. Note that $n=32$ from our base model and we set $s=48$ considering our hardware constraints and the efficiency of the scaled model, \textit{i.e.,} fitting between 7 and 13 billion parameters. Naturally, this leads to the removal of $m=8$ layers. The depthwise scaling process with $n=32, s=48,$ and $m=8$ is depicted in `Step 1: Depthwise Scaling' of Fig.~\ref{fig:dus}.

We note that a method in the community that also scale the model in the same manner~\footnote{\url{https://huggingface.co/Undi95/Mistral-11B-v0.1}} as `Step 1: Depthwise Scaling' of Fig.~\ref{fig:dus} has been concurrently developed.

\paragraph{Continued pretraining.}
The performance of the depthwise scaled model initially drops below that of the base LLM. Thus, we additionally apply the continued pretraining step as shown in `Step 2: Continued Pretraining' of Fig.~\ref{fig:dus}. Experimentally, we observe rapid performance recovery of the scaled model during continued pretraining, a phenomenon also observed in ~\citet{komatsuzaki2022sparse}. We consider that the particular way of depthwise scaling has isolated the heterogeneity in the scaled model which allowed for this fast performance recovery.

Delving deeper into the heterogeneity of the scaled model, a simpler alternative to depthwise scaling could be to just repeat its layers once more, \textit{i.e.,} from $n$ to $2n$ layers. Then, the `layer distance', or the difference in the layer indices in the base model, is only bigger than 1 where layers $n$ and $n+1$ are connected, \textit{i.e.,} at the seam.

However, this results in maximum layer distance at the seam, which may be too significant of a discrepancy for continued pretraining to quickly resolve.
Instead, depthwise scaling sacrifices the $2m$ middle layers, thereby reducing the discrepancy at the seam and making it easier for continued pretraining to quickly recover performance. We attribute the success of DUS to reducing such discrepancies in both the depthwise scaling and the continued pretraining steps. We also hypothesize that other methods of depthwise scaling could also work for DUS, as long as the discrepancy in the scaled model is sufficiently contained before the continued pretraining step.

\paragraph{Comparison to other up-scaling methods.}
Unlike ~\citet{komatsuzaki2022sparse}, depthwise scaled models do not require additional modules like gating networks or dynamic expert selection. Consequently, scaled models in DUS do not necessitate a distinct training framework for optimal training efficiency, nor do they require specialized CUDA kernels for fast inference. A DUS model can seamlessly integrate into existing training and inference frameworks while maintaining high efficiency.

\section{Training Details}
 After DUS, including continued pretraining, we perform fine-tuning of \pt in two stages: 1) instruction tuning and 2) alignment tuning.
\paragraph{Instruction tuning.} 
In the instruction tuning stage, the model is trained to follow instructions in a QA format~\cite{zhang2023instruction}. We mostly use open-source datasets but also synthesize a math  QA dataset to enhance the model's mathematical capabilities. A rundown of how we crafted the dataset is as follows.
First, seed math data are collected from the Math~\cite{hendrycks2021measuring} dataset only, to avoid contamination with commonly used benchmark datasets such as GSM8K~\cite{cobbe2021training}. Then, using a process similar to MetaMath~\cite{yu2023metamath}, we rephrase the questions and answers of the seed math data. We use the resulting rephrased question-answer pairs as a QA dataset and call it `Synth. Math-Instruct`.

\begin{table*}[t!]
\centering
\resizebox{0.85\linewidth}{!}{
\begin{tabular}{lccccccccc}
\toprule
Model & Size & Type  & H6 (Avg.) & ARC & HellaSwag & MMLU & TruthfulQA & Winogrande & GSM8K\\ \midrule
 \cellcolor{lp!60}\ft &  \cellcolor{lp!60}$\sim$ 11B&  \cellcolor{lp!60}Alignment-tuned& \cellcolor{lp!60}{\bf 74.20}  & \cellcolor{lp!60}{\bf 71.08} & \cellcolor{lp!60}88.16& \cellcolor{lp!60}66.21& \cellcolor{lp!60}{\bf 71.43} & \cellcolor{lp!60}83.58 &  \cellcolor{lp!60} 64.75\\ 
  Qwen 72B & $\sim$ 72B& Pretrained&73.60  &65.19 &85.94&{\bf 77.37}&60.19 &82.48 & {\bf70.43} \\
   Mixtral 8x7B-Instruct-v0.1 & $\sim$ 47B& Instruction-tuned&72.62  &70.22 &87.63&71.16&64.58 &81.37 & 60.73 \\
 % \cellcolor{lp!60}SOLAR-1-10.7B-SFT-Only* &  \cellcolor{lp!60}$\sim$ 11B&  \cellcolor{lp!60}Instruction-tuned& \cellcolor{lp!60} 71.11 & \cellcolor{lp!60}67.32 & \cellcolor{lp!60}85.96& \cellcolor{lp!60}65.95& \cellcolor{lp!60}58.80 & \cellcolor{lp!60} 82.08&  \cellcolor{lp!60}66.57 \\ 
    Yi 34B-200K & $\sim$ 34B& Pretrained&70.81  &65.36 &85.58&76.06&53.64 &82.56 & 61.64 \\
      % SOLAR-0-70B-16bit &  $\sim$ 70B&  Instruction-tuned& 70.11  & {\bf 71.08} & 87.89& 70.58&62.25 & 83.58 &  45.26 \\
    Yi 34B & $\sim$ 34B& Pretrained&69.42  &64.59 &85.69&76.35&56.23 &83.03 & 50.64 \\
  Mixtral 8x7B-v0.1 & $\sim$ 47B& Pretrained&68.42  &66.04 &86.49&71.82&46.78 &81.93 & 57.47 \\
  Llama 2 70B & $\sim$ 70B& Pretrained&67.87  &67.32 &87.33&69.83&44.92 &83.74 & 54.06 \\
         Falcon 180B & $\sim$ 180B& Pretrained&67.85  &69.45 &{\bf 88.86}&70.50&45.47 &{\bf86.90} & 45.94 \\
 \cellcolor{lp!60}\pt &  \cellcolor{lp!60}$\sim$ 11B&  \cellcolor{lp!60}Pretrained& \cellcolor{lp!60}66.04  & \cellcolor{lp!60}61.95 & \cellcolor{lp!60}84.60& \cellcolor{lp!60}65.48& \cellcolor{lp!60}45.04 & \cellcolor{lp!60}83.66 & \cellcolor{lp!60} 55.50 \\
Qwen 14B & $\sim$ 14B& Pretrained&65.86  &58.28 &83.99&67.70&49.43 &76.80 & 58.98 \\
Mistral 7B-Instruct-v0.2& $\sim$ 7B& Instruction-tuned&65.71  &63.14 &84.88&60.78&68.26 &77.19 & 40.03 \\
Yi 34B-Chat & $\sim$ 34B& Instruction-tuned&65.32  &65.44 &84.16&74.90&55.37 &80.11 & 31.92 \\
Mistral 7B& $\sim$ 7B& Pretrained&60.97  &59.98 &83.31&64.16&42.15 &78.37 & 37.83 \\
\bottomrule 
\end{tabular}
}
\caption{Evaluation results in the Open LLM Leaderboard for \pt and \ft along with other top-performing models. We report the scores for the six tasks mentioned in Sec.~\ref{para:eval} along with the H6 score (average of six tasks). We also report the size of the models in units of billions of parameters. The type indicates the training stage of the model and is chosen from \{Pretrained, Instruction-tuned, Alignment-tuned\}. Models based on \pt are colored purple. The best scores for H6 and the individual tasks are shown in bold.}
\label{tab:main_result}
\end{table*}

\paragraph{Alignment tuning.} In the alignment tuning stage, the instruction-tuned model is further fine-tuned to be more aligned with human or strong AI (\textit{e.g.,} GPT4~\cite{openai2023gpt4}) preferences using sDPO~\cite{kim2024sdpo}, an improved version of direct preference optimization (DPO)~\cite{rafailov2023direct}. Similar to the instruction tuning stage, we use mostly open-source datasets but also synthesize a math-focused alignment dataset utilizing the `Synth. Math-Instruct` dataset mentioned in the instruction tuning stage. 

The alignment data synthesis process is as follows. We take advantage of the fact that the rephrased question-answer pairs in Synth. Math-Instruct data are beneficial in enhancing the model's mathematical capabilities (see Sec.~\ref{sec:inst_abl}).
Thus, we speculate that the rephrased answer to the rephrased question is a better answer than the original answer, possibly due to the interim rephrasing step.
Consequently, we set the rephrased question as the prompt and use the rephrased answer as the chosen response and the original answer as the rejected response and create the \{prompt, chosen, rejected\} DPO tuple. We aggregate the tuples from the rephrased question-answer pairs and call the resulting dataset `Synth. Math-Alignment`.

\section{Results}
\subsection{Experimental Details}
\paragraph{Training datasets.}
We present details regarding our training datasets for the instruction and alignment tuning stages in Tab.~\ref{tab:datasets}.
We do not always use the entire dataset and instead subsample a set amount.
Note that most of our training data is open-source, and the undisclosed datasets can be substituted for open-source alternatives such as the MetaMathQA~\cite{yu2023metamath} dataset.

We reformatted the instruction datasets with an Alpaca-styled chat template. For datasets such as OpenOrca, which are derived from FLAN~\cite{longpre2023flan}, we filter data that overlaps with the benchmark datasets (see Tab.~\ref{tab:task_filter} in Appendix.~\ref{para:data_cont} for more information).
The alignment datasets are in the \{prompt, chosen, rejected\} triplet format.
We preprocess the alignment datasets following Zephyr~\cite{tunstall2023zephyr}.
We use Dataverse~\cite{park2024dataverse} for data preprocessing.

\paragraph{Evaluation.}
\label{para:eval}
In the HuggingFace Open LLM Leaderboard~\cite{open-llm-leaderboard}, six types of evaluation methods are presented: ARC~\cite{clark2018think}, HellaSWAG~\cite{zellers2019hellaswag}, MMLU~\cite{hendrycks2020measuring}, TruthfulQA~\cite{lin2022truthfulqa}, Winogrande~\cite{sakaguchi2021winogrande}, and GSM8K~\cite{cobbe2021training}. We utilize these datasets as benchmarks for evaluation and also report the average scores for the six tasks, \textit{e.g.,} H6.
We either submit directly to the Open LLM Leaderboard or utilize Evalverse~\cite{kim2024evalverse} for running evaluations locally.

\paragraph{Model merging.}
Model merging methods such as ~\citet{yadav2023ties} can boost model performance without further training.
We merge some of the models that we trained in both the instruction and alignment tuning stages.
We implement our own merging methods although popular open source also exist such as MergeKit\footnote{\url{https://github.com/cg123/mergekit}}.
\begin{table*}[t!]
\centering
\resizebox{0.90\linewidth}{!}{
\begin{tabular}{lcccccccccc}
\toprule
Model & Alpaca-GPT4 & OpenOrca & Synth. Math-Instruct& H6 (Avg.) & ARC & HellaSwag & MMLU & TruthfulQA & Winogrande & GSM8K\\ \midrule
 SFT v1 & {\color{ForestGreen} \cmark} & {\color{red} \xmark} & {\color{red} \xmark} &69.15 & {\bf67.66} & {\bf86.03}&65.88&{\bf60.12} &{\bf82.95} &52.24 \\
SFT v2 & {\color{ForestGreen} \cmark} & {\color{ForestGreen} \cmark} & {\color{red} \xmark} &69.21 &65.36  & 85.39&65.93& 58.47& 82.79& 57.32 \\
  SFT v3  & {\color{ForestGreen} \cmark} & {\color{ForestGreen} \cmark} & {\color{ForestGreen} \cmark} &70.03 &65.87  &85.55 &65.31& 57.93& 81.37&64.14 \\
   SFT v4 & {\color{ForestGreen} \cmark} & {\color{red} \xmark} & {\color{ForestGreen} \cmark} & 70.88& 67.32 &85.87 &65.87& 58.97&82.48 &64.75 \\
   SFT v3 + v4& {\color{ForestGreen} \cmark} & {\color{ForestGreen} \cmark} & {\color{ForestGreen} \cmark} &{\bf 71.11}&  67.32 &85.96 &{\bf 65.95}& 58.80& 82.08 &{\bf 66.57} \\
\bottomrule 
\end{tabular}
}
\caption{Ablation studies on the different datasets used for instruction tuning. `SFT v3+v4' indicates that the model is merged from `SFT v3' and `SFT v4' by simply averaging the model weights. The best scores for H6 and the individual tasks are shown in bold.}
\label{tab:sft_ablation}
\end{table*}

\begin{table*}[th!]
\centering
\resizebox{0.90\linewidth}{!}{
\begin{tabular}{lccccccccc}
\toprule
Model & Ultrafeedback Clean & Synth. Math-Alignment & H6 (Avg.) & ARC & HellaSwag & MMLU & TruthfulQA & Winogrande & GSM8K\\ \midrule
DPO v1 & {\color{ForestGreen} \cmark} & {\color{red} \xmark}& 73.06&71.42  & {\bf88.49}&{\bf66.14}& 72.04& 81.45&58.83 \\ 
DPO v2& {\color{ForestGreen} \cmark} & {\color{ForestGreen} \cmark}&{\bf 73.42} & {\bf 71.50} &88.28 &65.97& 71.71& {\bf 82.79}& {\bf 60.27}\\
DPO v1 + v2 & {\color{ForestGreen} \cmark} & {\color{ForestGreen} \cmark} & 73.21& 71.33 &88.36 &65.92& {\bf72.65}&{\bf82.79} &58.23 \\
\bottomrule 
\end{tabular}
}
\caption{Ablation studies on the different datasets used during the direct preference optimization (DPO) stage. `SFT v3' is used as the SFT base model for DPO. We name ablated models with the `DPO' prefix to indicate the alignment tuning stage. `DPO v1+v2' indicates that the model is merged from `DPO v1' and `DPO v2' by simply averaging the model weights. The best scores for H6 and the individual tasks are shown in bold.}
\label{tab:abl_dpo_data}
\end{table*}

\begin{table*}[th!]
\centering
\resizebox{0.75\linewidth}{!}{
\begin{tabular}{lcccccccc}
\toprule
Model & Base SFT Model& H6 (Avg.) & ARC & HellaSwag & MMLU & TruthfulQA & Winogrande & GSM8K\\ \midrule
DPO v2 & SFT v3 &73.42 & {\bf71.50} &{\bf88.28} &{\bf65.97}& 71.71& {\bf82.79}& 60.27\\
DPO v3 & SFT v3 + v4& {\bf 73.58}&  71.33&88.08 &65.39& {\bf 72.45}&81.93 &{\bf 62.32} \\
\bottomrule 
\end{tabular}
}
\caption{Ablation studies on the different SFT base models used during the direct preference optimization (DPO) stage. Ultrafeedback Clean and Synth. Math-Alignment datasets are used. We name ablated models with the `DPO' prefix to indicate the alignment tuning stage. The best scores for H6 and the individual tasks are shown in bold.}
\label{tab:dpo_sft_base}
\end{table*}

\begin{table*}
\centering
\resizebox{0.65\linewidth}{!}{
\begin{tabular}{lccccccc}
\toprule
Model &  H6 (Avg.) & ARC & HellaSwag & MMLU & TruthfulQA & Winogrande & GSM8K\\ \midrule
Cand. 1 & {\bf73.73}& 70.48 &87.47 &65.73& 70.62&81.53 &{\bf66.57} \\
Cand. 2& 73.28 & {\bf71.59} &{\bf88.39} &{\bf66.14}&{\bf72.50} & {\bf81.99}&59.14 \\ 
\bottomrule
\end{tabular}
}
\caption{Performance comparison amongst the merge candidates. `Cand. 1' and `Cand. 2' are trained using the same setting as `DPO v2' and `DPO v3', respectively, but with slightly different hyper-parameters. The best scores for H6 and the individual tasks are shown in bold.}
\label{tab:merge_cand}
\end{table*}

\subsection{Main Results}
We present evaluation results for our \pt and \ft models along with other top-performing models in Tab.~\ref{tab:main_result}.
\pt outperforms other pretrained models of similar sizes, such as Qwen 14B and Mistral 7B, which shows that DUS is an effective method to up-scale base LLMs. Furthermore, despite the smaller size, \ft scores the highest in terms of H6, even surpassing the recent top-performing open-source LLM Mixtral 8x7B-Instruct-v0.1 or Qwen 72B. The above results indicate DUS can up-scale models that are capable of achieving state-of-the-art performance when fine-tuned. We also report data contamination results for \ft in Appendix~\ref{para:data_cont}.

\subsection{Ablation Studies}
We present ablation studies for both the instruction and alignment tuning stages.
Note that the evaluation results for the following studies are ran locally and may vary from results obtained by submitting to the Open LLM Leaderboard.

\subsubsection{Instruction Tuning}
\label{sec:inst_abl}
\paragraph{Ablation on the training datasets.}
We present ablation studies using different training datasets for the instruction tuning in Tab.~\ref{tab:sft_ablation}.
The ablated models are prefixed with SFT for supervised fine-tuning.
`SFT v1' only uses the Alpaca-GPT4 dataset, whereas `SFT v2' also uses the OpenOrca dataset.
`SFT v3' uses the Synth. Math-Instruct dataset along with the datasets used in `SFT v2'.
Similarly, `SFT v4' uses the Synth. Math-Instruct dataset along with the datasets used in `SFT v1'.

First, we analyze how Alpaca-GPT4 and OpenOrca affect the trained models.
The first ablated model, `SFT v1', which used only the Alpaca-GPT4 dataset for training, resulted in $69.15$ for H6. When we add the OpenOrca dataset to train the second ablated model, `SFT v2', the resulting H6 score is $69.21$, which is little change from $69.15$ of `SFT v1'. However, the task scores vary more as `SFT v2' gets a substantially higher GSM8K score of $57.32$ compared to $52.24$ of `SFT v1' but also gets noticeably lower scores across the board for ARC, HellaSwag, and TruthfulQA.
This seems to indicate that using OpenOrca results in a model that behaves differently from using only Alpaca-GPT4.

Second, we investigate whether Synth. Math-Instruct dataset is beneficial. For `SFT v3', we add the Synth. Math-Instruct dataset, which boosts GSM8K scores to $64.14$ and achieves comparable scores for the other tasks.
Interestingly, when we add the Synth. Math-Instruct dataset to `SFT v1' to train `SFT v4', we get our highest H6 score of $70.88$ with higher scores than `SFT v3' for all tasks. From the above, we can see that adding the Synth. Math-Instruct dataset is helpful.

Lastly, we see whether merging models trained with and without OpenOrca can boost performance. In the first analysis, we saw that using OpenOrca resulted in a model that behaved differently from the model that was trained without OpenOrca. Building on this intuition, we merge `SFT v3' and `SFT v4' as they are the best-performing models with and without OpenOrca.
To our surprise, the resulting merged model `SFT v3+v4' retains the high scores for non-GSM8K tasks from `SFT v4' but also achieves a higher GSM8K score than `SFT v3' or `SFT v4'. Thus, we see that merging models that specialize in different tasks is a promising way to obtain a model that performs well generally.

\begin{table*}
\centering
\resizebox{0.75\linewidth}{!}{
\begin{tabular}{lcccccccc}
\toprule
Model & Merge Method& H6 (Avg.) & ARC & HellaSwag & MMLU & TruthfulQA & Winogrande & GSM8K\\ \midrule
Merge v1 & Average (0.5, 0.5) & 74.00& {\bf71.16} &88.01 &66.14& 71.71&{\bf82.08} &64.90 \\
Merge v2 & Average (0.4, 0.6) &73.93 & 71.08 &{\bf88.08} &{\bf66.27}&{\bf71.89} & 81.77&64.52 \\ 
 Merge v3 & Average (0.6, 0.4) & {\bf 74.05}& 71.08 &87.88 &66.13&  71.61& {\bf 82.08}& {\bf 65.50} \\
Merge v4 & SLERP &73.96 &{\bf71.16}  &88.03 &66.25& 71.79&81.93 &64.59 \\ 
\bottomrule
\end{tabular}
}
\caption{Ablation studies on the different merge methods used for obtaining the final model. We use `Cand. 1' and `Cand. 2' from Tab.~\ref{tab:merge_cand} as our two models for merging. We name the merged models with the `Merge' prefix to indicate they are merged. The best scores for H6 and the individual tasks are shown in bold.}
\label{tab:merge_abl}
\end{table*}

\subsubsection{Alignment Tuning}
As we utilize sDPO for practical alignment tuning, there are additional aspects to ablate such as the SFT base models used. Thus, we present ablations for the different training datasets used for training, the different SFT base models to initialize the sDPO training, and finally, the model merging strategy to obtain the final alignment-tuned model.

\paragraph{Ablation on the training datasets.}
We ablate on the different alignment datasets used during DPO in Tab.~\ref{tab:abl_dpo_data}. 
We use `SFT v3' as the SFT base model for DPO.
`DPO v1' only uses the Ultrafeedback Clean dataset while `DPO v2' also used the Synth. Math-Alignment dataset.

First, we test how Ultrafeedback Clean and Synth. Math-Alignment impacts model performance. For `DPO v1', it achieves $73.06$ in H6, which is a substantial boost from the SFT base model score of $70.03$. However, we note that while scores for tasks like ARC, HellaSwag, and TruthfulQA all improved by good margins, the score for GSM8K is $58.83$, which is lower than the SFT base model score of $64.14$. Adding Synth. Math-Alignment to train `DPO v2', we see that the GSM8k score improves to $60.27$, which is lower than the SFT base model but still higher than `DPO v1'. Other task scores are also not negatively impacted by adding Synth. Math-Alignment. Thus, we can conclude that adding Synth. Math-Alignment is beneficial for H6.

Then, we experiment whether merging `DPO v1' and `DPO v2' is beneficial. Unfortunately, `DPO v1+v2' scores $73.21$ in H6, which is worse than `DPO v2'. More importantly, the gain in the GSM8K score from adding Synth. Math-Alignment is gone, which is undesirable. One reason for this could be that `DPO v2' is a strict improvement over `DPO v1', unlike the case for merging `SFT v3' and `SFT v4' where the models had different strengths and weaknesses.

\paragraph{Ablation on the SFT base models.}
When applying DPO, we start from a model that is already instruction tuned \textit{,i.e.,} the SFT base model and ablate on using different SFT base models. We use Ultrafeedback Clean and Synth. Math-Alignment datasets for this ablation. Each of the ablated models is trained as follows.
`DPO v2' uses `SFT v3' as the base SFT model, while `DPO v3' uses `SFT v3+v4' as the SFT base model instead.

Note that `SFT v3+v4' has higher scores on all tasks compared to `SFT v3', and the gap is especially large for ARC ($+1.45$) and GSM8K ($+2.43$).
Surprisingly, the two models perform similarly in terms of H6. A closer look at the scores for the individual tasks shows only a small margin in the GSM8K scores, and other task scores show little difference. Thus, the performance gaps in certain tasks in the SFT base models do not always carry over to the alignment-tuned models.

\paragraph{Ablation on different merge methods.}
From Tab.~\ref{tab:sft_ablation}, we saw that merging two models that have different strengths can be beneficial to performance. To utilize this for the alignment-tuned model as well, we train two models named `Cand. 1' and `Cand. 2' using the same training dataset and SFT base model as `DPO v2' and `DPO v3' but with different hyper-parameters to maximize each model's respective strengths. We compare `Cand. 1' and `Cand. 2' in Tab.~\ref{tab:merge_cand} where we can see that `Cand. 1' has high GSM8K scores but relatively low scores for the other tasks, whereas `Cand. 2' has low scores for GSM8K but high scores for the other tasks. We merge these two models using various methods and ablate the results in Tab..~\ref{tab:merge_abl}.

We use two merge methods: 1) Average ($a$, $b$), where a and b denote the weighting for `Cand. 1' and `Cand. 2' when averaging weights and 2) SLERP~\cite{shoemake1985animating}.
We use ($0.5$, $0.5$), ($0.4$, $0.6$), and ($0.6$, $0.4$) for Average ($a$, $b$).
From Tab.~\ref{tab:merge_abl}, we can see that the different merge methods have little effect on the H6 scores. The scores for the individual tasks also do not differ by much, suggesting that as long as the merge candidates have sufficiently different strengths, the exact merge method may not be as crucial. Thus, we chose `Merge v1' as our \ft model.

\section{Conclusion}
We introduce \pt and its fine-tuned variant \ft, which are depth up-scaled (DUS) models with 10.7 billion parameters\footnote{Preprint version is available on \url{https://arxiv.org/abs/2312.15166}.}. They show superior performance over models like Llama 2, Mistral 7B, and Mixtral-7B-Instruct in essential NLP tasks while maintaining computational efficiency. Thus, DUS is effective in scaling-up highly performant LLMs from smaller ones. With more exploration, DUS could be further improved, paving a new path to efficiently scaling LLMs.

\section*{Acknowledgements}
We would like to extend our gratitude to the teams at Hugging Face, particularly Clémentine Fourrier, Lewis Tunstall, Omar Sanseviero, and Philipp Schmid. Our appreciation also extends to the teams at AWS, notably Rahul Sharma, Jeongwon Yoon, Nieves Garcia, Ritesh Vajaria, Gal Oshri, Jay Kwon, Brandon Lee and Effie Bae. We are grateful to the teams at Korea Telecom (KT), especially Jin Hyoung Lee, Jungsuk Park, Sungjoon Park, Hong-rae Wang, Kyeongsoo Jung, and Sunyoong Yoon, whose significant support has been instrumental in ensuring the broad compatibility of our model. Additionally, we would like to extend our thanks to the open community for their invaluable contributions and feedback.

\section*{Limitations}
Our study on the Depth Up-Scaling (DUS) has important limitations and considerations. One key limitation is the need for more thorough explorations of hyperparameters used in the DUS approach. Namely, we removed $m=8$ layers from both ends of our base model, primarily due to hardware limitations. However, we have not yet determined if this value is optimal for enhancing performance. The extended time and cost of continued pretraining made it challenging to conduct more comprehensive experiments, which we aim to address in future work through various comparative analyses.

In terms of the model's broader implications, there are several points to note. The model's significant computational demands for training and inference might limit its use, especially for those with restricted computational resources. Additionally, like all machine learning models, it is vulnerable to biases in its training data, which could lead to skewed outcomes in certain situations. Furthermore, the substantial energy consumption required for training and operating the model raises environmental concerns, which are critical in the pursuit of sustainable AI development.

Lastly, while the fine-tuned variant of the model shows improved performance in following instructions, it still requires task-specific fine-tuning for optimal performance in specialized applications. This fine-tuning process can be resource-intensive and not always effective. Recognizing and addressing these limitations is essential for a comprehensive understanding of the proposed Large Language Model's capabilities and for guiding future research and development in the field of LLMs.

\section*{Ethics Statement}
We conscientiously address and emphasize the commitment of \pt in maintaining the highest ethical standards. First, we highlight that \ft has shown low levels of data contamination in our evaluations, a testament to our rigorous data handling and processing protocols. This aspect is crucial, as it underpins the reliability and integrity of the results obtained from SOLAR.

Furthermore, during the course of our experiments, we ensured that all setups and methodologies employed steer clear of any potential ethical pitfalls. This preemptive consideration and avoidance of ethically questionable practices underscore our dedication to conducting research that is not only innovative but also responsible.

Additionally, we ensure that SOLAR complies with general ethical considerations in all aspects of its operation. This includes adherence to privacy norms, respect for intellectual property, and ensuring the absence of bias in our algorithms. Our commitment to these ethical principles is unwavering, and we believe it significantly contributes to the credibility and societal acceptance of SOLAR.

In conclusion, the ethical framework within which SOLAR operates is robust and comprehensive, ensuring that our advancements in this field are not only scientifically sound but also ethically responsible.

\bibliography{anthology,custom}
\bibliographystyle{acl_natbib}

\clearpage
\appendix

\section{Contributions}

The contributions of this study are as follows:
\begin{itemize}
    \item \textbf{Introduction of the SOLAR 10.7 Billion-Parameter Model}: We have released the SOLAR 10.7B model, which is not only depthwise scaled but also continually pretrained. The availability of \pt under the Apache 2.0 license permits commercial usage, enabling the integration of this advanced model into a diverse range of products and services. This bridges the gap between academic research and practical applications, fostering wider accessibility and utility in various fields.
    
    \item \textbf{Superior Performance Across Diverse Benchmarks}: \pt excels in various benchmarks, outperforming established models like Llama 2 and Mistral 7B in reasoning, mathematics, and the MMLU framework.
    
    \item \textbf{Advancement in Instruction-Following Capabilities}: The introduction of \ft, a variant fine-tuned for enhanced instruction-following abilities, marks a significant improvement in the model's ability to understand and execute complex instructions.
\end{itemize}

Sanghoon Kim, Dahyun Kim, Chanjun Park, Wonsung Lee, Wonho Song, Yunsu Kim and Hyeonwoo Kim contributed equally to this paper. Sanghoon Kim led the Foundation Model part, with Dahyun Kim, Wonho Song, Yunsu Kim, and Hyeonwoo Kim. Chanjun Park led the Data and Evaluation (Data-Centric LLM) part, with Yungi Kim, Jihoo Kim, Changbae Ahn, Seonghoon Yang, Sukyung Lee, and Hyunbyung Park. Wonsung Lee led the Adaptation Modeling part, with Gyoungjin Gim, Hyeonju Lee, and Mikyoung Cha. Hwalsuk Lee performed the role of the overall project operation. Dahyun Kim and Chanjun Park were the main technical writers. All these individuals contributed to the creation of \pt. 

\section{Related Works and Background}
\subsection{Large Language Models} Following the advent of context-based language models, various studies have revealed a “scaling law”~\cite{kaplan2020scaling, hernandez2021scaling, anil2023palm}, demonstrating a positive correlation between the size of model and training data and model performance. This has led to the emergence of Large Language Models (LLMs). Unlike previous language models, LLMs possess the ability for In-context learning, including Zero-shot learning~\cite{radford2019language} and Few-shot learning~\cite{brown2020language}, allowing them to perform new tasks without updating model weights. These capabilities of LLMs, not evident in smaller models, are referred to as Emergent abilities~\cite{wei2022emergent}.

\subsection{Mixture of Experts}
In the landscape of machine learning architectures, the Mixture of Experts (MoE) models like ~\cite{shazeer2017outrageously, shen2019mixture, komatsuzaki2022sparse} has gained attention for its capability to address the challenges posed by complex and heterogeneous data. MoE models offer notable benefits, including enhanced output diversity, allowing for the capture of intricate patterns within the input space. Moreover, their computational efficiency, especially when implemented in a sparse form, has made them valuable in scenarios where resource constraints are a consideration~\cite{shazeer2017outrageously, komatsuzaki2022sparse}.

However, efficient implementation of MoE models poses a considerable challenge, primarily due to the intricacies associated with dynamic routing and load-imbalanced computation~\cite{gale2023megablocks}. Existing hardware and software for deep learning, such as TPUs and XLA compilers, often demand static knowledge of tensor shapes, making MoE implementation on TPU challenging. 

While GPU implementation offers more flexibility, sparse computation compatibility becomes a hurdle. Striking the right balance between fixing the size of each expert to facilitate efficient computation and maintaining model quality creates a tradeoff between information preservation and hardware efficiency. This tradeoff, in turn, necessitates careful consideration during hyperparameter tuning, adding a layer of complexity to the implementation of MoE models, potentially offsetting their advantages. Given the formidable challenges in MoE model implementation, it becomes almost inevitable for researchers and practitioners to resort to specialized tools and frameworks, such as Tutel~\cite{hwang2023tutel} or Megablocks~\cite{gale2023megablocks}.

Departing from the horizontal expansion characteristic of MoE models, the DUS method introduces model scaling in the vertical dimension. Notably, DUS does not introduce dynamism in the scaled model, which significantly reduces the complexity when compared to MoE. This shift in approach offers a unique and more straightforward way of working, moving away from conventional MoE challenges.
Not only that, DUS also undergoes continued pretraining to quickly recover performance of the scaled model.

\subsection{Prompt Engineering}
A key research area to harness the emergent abilities of LLMs is prompt engineering. Prompt engineering is the study of how to design inputs (prompts) that enable LLMs to better perform specific tasks. A prime example of this research is Chain-of-Thought (CoT)~\cite{wei2022chain}, which proposes CoT prompting that decomposes multi-step problems into a series of intermediate reasoning steps. Moreover, efforts are underway to replace even such prompt engineering with LLMs~\cite{yang2023large}.

\subsection{Instruction Tuning}
To enhance the steerability of LLMs, instruction tuning~\cite{wei2021finetuned}  has emerged as a learning technique. This involves fine-tuning LLMs using data formatted as (instruction, input, output) for various tasks~\cite{wang2022self}. Instruction tuning allows for targeted adjustments, providing a more controlled and task-oriented improvement to the model's capabilities.

Before instruction tuning, existing methods faced challenges in effectively guiding and controlling the behavior of large language models~\cite{zhang2023instruction}. The sheer complexity of these models made it difficult to ensure precise and task-oriented responses. The need for a more targeted approach arose from the limitations of existing methods, leading to the development of instruction tuning.
This targeted approach enables better control over the model's behavior, making it more suitable for specific tasks and improving its overall performance in alignment with user-defined objectives. Therefore, instruction tuning is computationally efficient and facilitates the rapid adaptation of LLMs to a specific domain without requiring extensive retraining or architectural changes.

\subsection{Alignment Tuning}
LLM has been observed to generate sentences that may be perceived as linguistically incongruent by human readers since they learned not human intention, but only vast knowledge across various domains in the pretraining step~\cite{ziegler2019fine}. To overcome this limitation and align with human intentions, previous research~\cite{ziegler2019fine} have proposed Reinforcement Learning with Human Feedback (RLHF). RLHF operates by learning a reward model based on human preferences, employing reinforcement learning to guide the LLM towards prioritizing answers with the highest reward scores. This process enhances the safety, propriety, and overall quality of the generated responses. Despite demonstrating satisfactory performance, RLHF encounters challenges such as managing numerous hyperparameters and necessitating the incorporation of multiple models (policy, value, reward, and reference models).

In response to these challenges, the supervised fine-tuning based approaches have proposed, such as Rank Responses to align Human Feedback (RRHF)~\cite{yuan2023rrhf}, Reward rAnked FineTuning (RAFT)~\cite{dong2023raft}, and Direct Policy Optimization (DPO)~\cite{intel2023orcadpo}. They avoid the complexities associated with reinforcement learning while achieving empirical performance comparable to RLHF.  Among them, DPO that we used directly guides the LLM to increase the probability of positive responses and decrease the probability of negative responses through a "direct" approach. Interestingly, DPO demonstrates more stable learning results compared to RLHF, despite its simple training approach.

\subsection{Data Contamination}
Recent researches~\cite{zhou2023don, sainz2023nlp, golchin2023time, deng2023investigating} emphasize the need to measure whether a specific benchmark was used to train the large language models. There are three types of the data contamination: guideline, raw text and annotation~\cite{sainz2023nlp}. \textbf{Guideline contamination} occurs when a model accesses detailed annotation guidelines for a dataset, providing advantages in specific tasks, and its impact should be considered, especially in zero and few-shot evaluations. \textbf{Raw text contamination} occurs when a model has access to the original text. Wikipedia is widely used as a pretraining data, but also as a source for creating new datasets. The caution is advised in the development of automatically annotated datasets sourced from the web. \textbf{Annotation contamination} occurs when the annotations of the specific benchmark are exposed during model training. 

\section{Additional Information}
We present additional information for the sake of space in the main paper.

\paragraph{Filtered task names.}
We present task names we use to filter FLAN dervied datasets such as OpenOrca in Table~\ref{tab:task_filter}.

\begin{table}[h!]
\resizebox{0.8\columnwidth}{!}{
\begin{tabular}{l}
\toprule
Filtered Task Name \\ \midrule
task228\_arc\_answer\_generation\_easy\\
ai2\_arc\/ARC\-Challenge:1.0.0\\
ai2\_arc\/ARC\-Easy:1.0.0\\
task229\_arc\_answer\_generation\_hard\\
hellaswag:1.1.0\\
task1389\_hellaswag\_completion\\
cot\_gsm8k\\
cot\_gsm8k\_ii\\
drop:2.0.0\\
winogrande:1.1.0\\
\bottomrule 
\end{tabular}
}
\caption{Task names that we use to filter data for FLAN derived datasets such as OpenOrca.}
\label{tab:task_filter}
\end{table}

\begin{table}[th!]
\centering
\resizebox{1.00\linewidth}{!}{
\begin{tabular}{ccccccc}
\toprule
ARC & HellaSwag & MMLU & TruthfulQA & Winogrande & GSM8K\\ \midrule
0.06 & N/A&0.15& 0.28& N/A&0.70 \\ 
\bottomrule 
\end{tabular}
}
\caption{Data contamination test results for \ft. We show `result < 0.1, \%` values where a value higher than 0.9 indicates high probability of data contamination. HellaSwag and Winogrande datasets are not currently supported. We set \pt as our reference model when performing the data contamination tests.}
\label{tab:data_contam}
\end{table}

\paragraph{Results on data contamination.}
\label{para:data_cont}
To show the integrity of \ft, we also report the data contamination test~\cite{shi2023detecting} results in Table.~\ref{tab:data_contam}. All four tested benchmark datasets yield results well below the contamination threshold, affirming the absence of data contamination in our model. One interesting point is that the value for GSM8K is noticeably higher than for other datasets, even without contamination.
One potential reason for this is the stronger data similarity in math-related instruction datasets.

\end{document}